\documentclass[letterpaper, 10 pt, conference]{ieeeconf}  

\IEEEoverridecommandlockouts                              

\pdfoutput=1
\usepackage{graphics} 
\usepackage{epsfig} 
\usepackage{amsmath} 
\usepackage{bm}
\usepackage[table]{xcolor}
\usepackage{color}
\usepackage{subfigure}
\usepackage{subcaption}
\usepackage{multirow}
\usepackage{stfloats}
\usepackage{cite}

\title{\LARGE \bf
A Visual-Inertial Motion Prior SLAM for Dynamic Environments
}
\author{Weilong Sun$^{1}$, Yumin Zhang$^{2*}$ and Boren Wei$^{2}$
\thanks{$^{1}$Weilong Sun, $^{2}$Yumin Zhang and $^{2}$Boren Wei are with the College of Instrumental Science and Optoelectronic Engineering, Beijing University of Aeronautics and Astronautics, Beijing 100191, China.}%
\thanks{* Corresponding author: Prof. Yumin Zhang.}%
\thanks{This research would like to thank colleges of the MGNC Lab. , Beijing University of Aeronautics and Astronautics and  Prof. Wei Sheng for useful discussions.}%
}

\begin{document}

\maketitle
\thispagestyle{empty}
\pagestyle{empty}

\begin{abstract}

The Visual-Inertial Simultaneous Localization and Mapping (VI-SLAM) algorithms which are mostly based on static assumption are widely used in fields such as robotics, UAVs, VR, and autonomous driving. To overcome the localization risks caused by dynamic landmarks in most VI-SLAM systems, a robust visual-inertial motion prior SLAM system, named (\textit{IDY-VINS}), is proposed in this paper which effectively handles dynamic landmarks using inertial motion prior for dynamic environments to varying degrees. Specifically, potential dynamic landmarks are preprocessed during the feature tracking phase by the probabilistic model of landmarks' minimum projection errors which are obtained from inertial motion prior and epipolar constraint. Subsequently, a robust and self-adaptive bundle adjustment residual is proposed considering the minimum projection error prior for dynamic candidate landmarks. This residual is integrated into a sliding window based nonlinear optimization process to estimate camera poses, IMU states and landmark positions while minimizing the impact of dynamic candidate landmarks that deviate from the motion prior. Finally, \textcolor{black}{a clean point cloud map without `ghosting effect' is obtained that contains only static landmarks}. Experimental results demonstrate that our proposed system outperforms state-of-the-art methods in terms of localization accuracy and time cost by robustly mitigating the influence of dynamic landmarks.
\end{abstract}

\section{INTRODUCTION}
In environments without satellite positioning, V-SLAM obtains information of the surrounding environment through visual sensors to solve its own motion. It has the advantages of low cost, high precision, no need for any environmental prior, and rich applicable scenarios. However, visual-only sensor is easily affected by external environmental and motion factors, such as dynamic landmarks, the lack of texture, lighting conditions, motion blur, and few common field of view (FOV) \cite{r1}. Therefore, an inertial measurement unit (IMU), which can provide relatively accurate pose estimation information without being affected by the environment in a short period of time, can be integrated to compensate for the deficiencies of visual-only sensor \cite{imu_visual}.

\textcolor{black}{However, the assumption of static landmarks in visual-related SLAM still makes it have hidden localization risk in environments with dynamic landmarks \cite{dy_problem}.} The dynamic landmarks will affect the accuracy of visual positioning, and in extreme cases, they will lead to rapid deterioration of positioning. \textcolor{black}{At the same time, a clean static feature map is also of great significance for Life-Long SLAM systems. This is because that the `ghosting effect' caused by the dynamic landmarks in the feature map will greatly affect the localization based on this prior map.}

\begin{figure}[!h]
\centering
\includegraphics[scale=0.20]{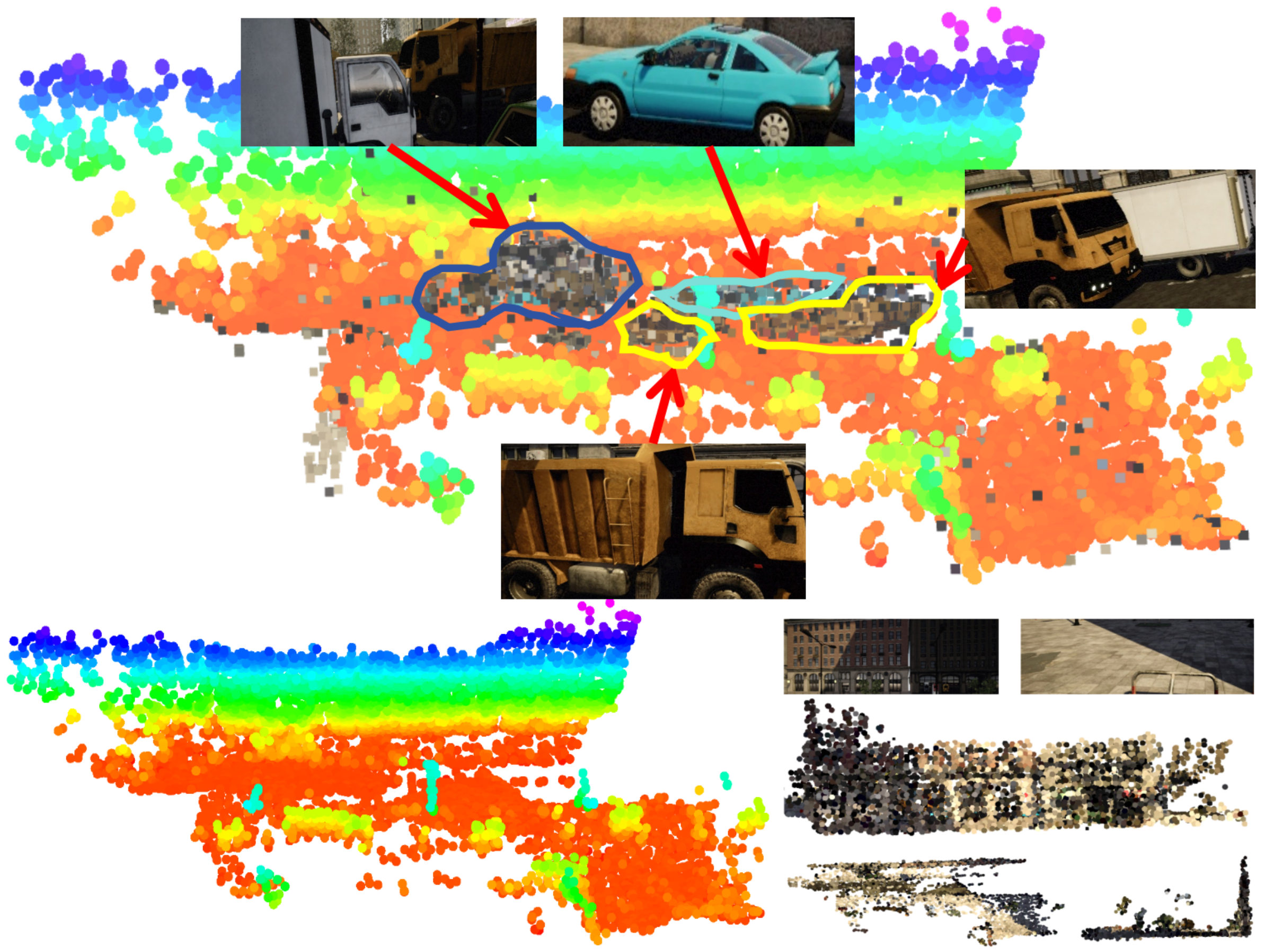}
\caption{\textbf{Point cloud map:} \textcolor{black}{The upper part is a point cloud map containing both static and dynamic landmarks which include the dynamic landmarks that were excluded during the feature tracking phase with triangulation, as well as the dynamic landmarks that were confirmed after the nonlinear optimization process. And the lower part is a clean map containing only static landmarks.}}
\label{figurelabel}
\end{figure}

In recent years, research on visual-related SLAM in dynamic environments has gradually become a focus. Most studies seek solutions through deep learning methods such as semantic recognition \cite{semantic1,semantic2} while some studies use geometric methods \cite{multiple_v_geometry,8POINTS}. However, deep learning methods are always limited by scenes and training processes while geometric methods always have low-precision and unstable performance.

In order to solve the problem of dynamic landmarks, a robust visual-inertial motion prior SLAM system is proposed in this paper which is applicable to a large range of dynamic environments and only uses the motion prior propagated by the IMU in a short period of time to process dynamic landmarks. This paper presents our work from the following aspects:

(1) \textbf{The preprocessing of potential dynamic landmarks with inertial motion prior:} The minimum projection errors of landmarks are obtained from inertial motion prior and epipolar constraint to eliminate dynamic landmarks that significantly deviate from the motion prior and mark dynamic candidate landmarks during the feature tracking phase.

(2) \textbf{The robust and self-adaptive bundle adjustment residual for dynamic candidate landmarks:} A dynamic visual residual is constructed \textcolor{black}{which considers the minimum projection error prioris and is used in nonlinear optimization process to reduce the impact of dynamic candidate landmarks that deviate from the motion prior.}

Compared with existing methods in the VIODE dataset \cite{viode} which contains dynamic objects within different ranges, superior robust and stable localization accuracy, and good real-time performance of our approach are demonstrated. \textcolor{black}{And a clean point cloud map without `ghosting effect' is obtained after the post-processing.} In addition, the localization accuracy is compared in the EUROC dataset \cite{euroc}, and the result demonstrates stable and superior performance of our approach in static environments. 
\begin{figure*}[htpb]
\centering
\includegraphics[scale=0.3]{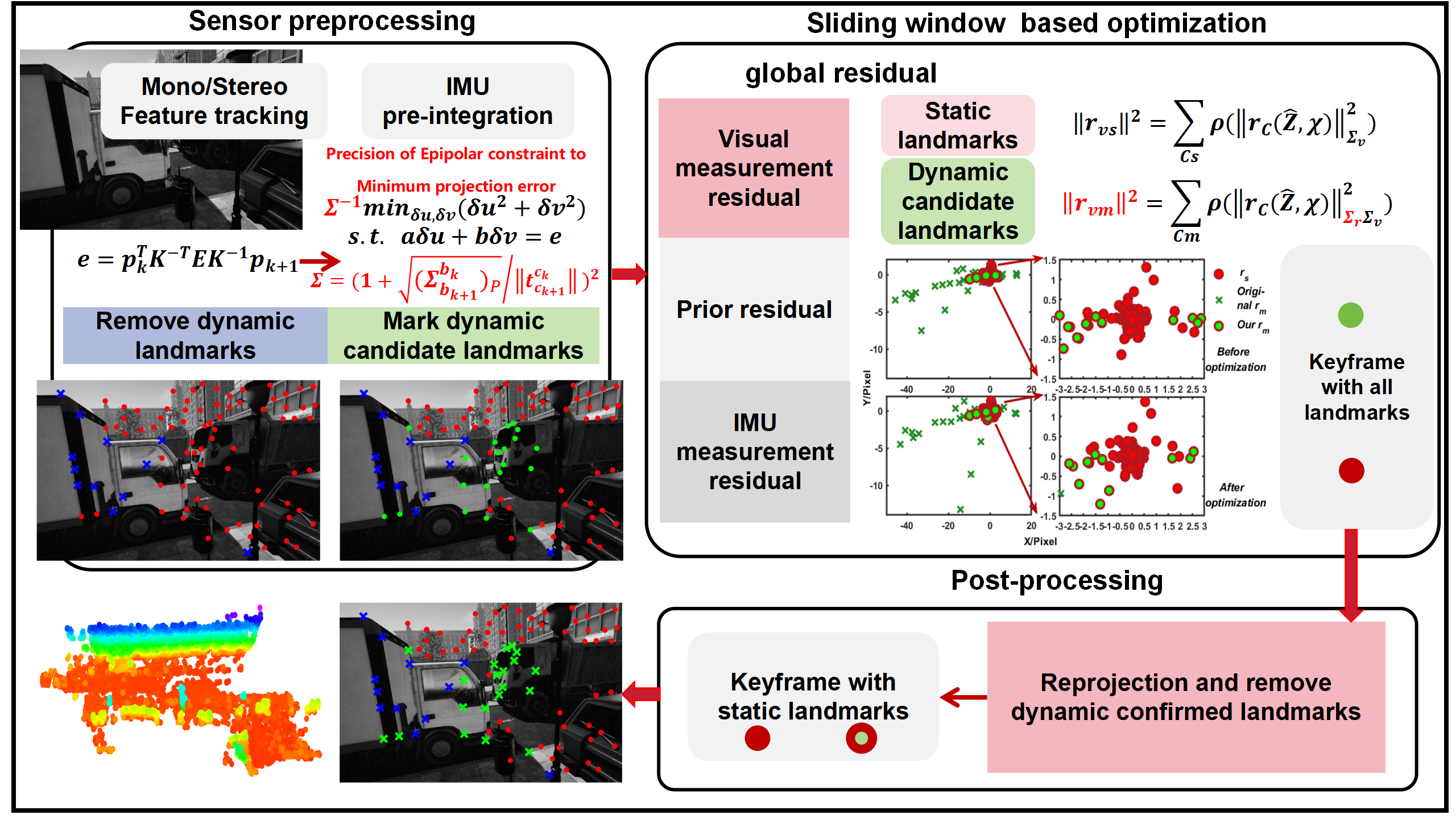}
\caption{\textcolor{black}{\textbf{System overview:}} Our system consists of two main components: (1) Sensor preprocessing, which gets landmarks' minimum projection errors from inertial motion prior and epipolar constraint to eliminates dynamic landmarks and mark dynamic candidate landmarks; (2) Sliding window based optimization with \textcolor{black}{robust and self-adaptive} bundle adjustment residual, which minimizes the impact of dynamic candidate landmarks that deviate from the motion prior. And the post-processing component, which removes confirmed dynamic landmarks from sliding window.}
\label{figurelabel}
\end{figure*}
\section{RELATED WORK}
VI-SLAM algorithms utilize the inertial measurement unit to correct the scale and the camera poses while V-SLAM algorithms only use visual sensors. Generally, they can be divided into filter-based methods and optimization-based methods which have become mainstream. For example, the ROVIO \cite{ROVIO} employed the extended Kalman filter (EKF) method, while OKVIS \cite{OKVIS} proposed a keyframe-based framework using optimization process. The ORB descriptor \cite{orb-slam} for feature matching was employed to optimize camera poses and landmark positions with IMU measurements in ORB-SLAM3 \cite{orb-slam3}. The VINS-Fusion \cite{vins_fusion} is an extended version of VINS-Mono \cite{vins_mono}, which supports stereo cameras with IMU and adopts feature tracking instead of descriptor matching. However, the above VI-SLAM methods still have potential limitations in dealing with dominant dynamic objects.

Recently, many researchers have proposed a variety of methods for dealing with dynamic objects in the V-SLAM and VI-SLAM algorithms.
\subsection{Methods based on vision geometry}
Fan et al. \cite{fan_geome} proposed a multi-view geometry-based method using RGB-D camera. The type of each landmark is determined to be dynamic or static according to the geometric relationship between the camera motion and the landmark's states after obtaining the camera pose by minimizing the reprojection error. The RANSAC method \cite{ransac,ransac2} can also be used to remove dynamic landmarks in small-scale dynamic scenes. In fact, it can distinguish between inliers and outliers in various anomalies such as noise, occlusion and dynamic objects, improving the robustness of model parameter estimation. The multi-view geometry-based methods assume that the camera pose estimation is accurate enough, which will lead to the failure of the algorithms when the estimation is inaccurate due to the presence of too many dynamic objects.
\subsection{Methods based on deep learning}
In DynaSLAM \cite{dynaslam}, the dynamic landmarks in masked areas where predefined dynamic objects by deep learning networks are removed, while the types of remaining landmarks are determined through multi-view geometry. Wen et al. \cite{dynamic_slam} proposed a method that combines semantic segmentation information and spatial motion information of associated pixels to deal with dynamic objects. Masoud S. Bahraini et al. \cite{deep_ransac} adopted a method that combines multi-level RANSAC operators with deep learning to distinguish between dynamic landmarks and static landmarks. Although deep learning methods can successfully eliminate the landmarks of dynamic objects, there are still some problems, such as: many methods require predefined dynamic objects; when only a part of a dynamic object can be seen due to occlusion, it may not be detected; the detected dynamic objects may not be in movement but are temporarily stationary.

\subsection{Methods based on inertial sensor}
Cui et al. \cite{VIMU} detected and removed dynamic landmarks according to the epipolar constraint provided by the prior of IMU pre-integration, and conducted states estimation by utilizing more accurate static landmarks. The DynaVINS \cite{dynavins} proposed a robust bundle adjustment system based on the IMU pre-integration regularization factor and the momentum factor to reject dynamic landmarks. It needs to perform optimization based on the regularization factor and the momentum factor before formal nonlinear optimization to obtain the weight corresponding to each landmark in order to reject dynamic landmarks which leads to the fact that it cannot operate in real time. At the same time, its method may not be suitable for other challenging scenarios in visual SLAM, such as changes in lighting and drastic movements of camera. The method in this paper is also based on the motion prior propagated by the IMU which performs well both in dynamic environments and static environments with other challenging scenarios, and has better real-time performance.

\section{PREPROCESSING}

In this section, the method for preprocessing potential dynamic landmarks with inertial motion prior is introduced. Due to the IMU propagation motion prior between two frames of camera, the minimum projection error is obtained through the landmark's precision of the epipolar constraint, then its probability model is modeled and applied to determine dynamic landmarks will be eliminated and dynamic candidate landmarks will be marked during the tracking phase.

\subsection{Inertial Motion Piror} The relative translation and rotation in the camera coordinate system can be calculated by IMU propagation result and the extrinsic parameters between IMU and camera. The residual covariance matrix of IMU pre-integration \cite{imu_preintergration} can be linearly propagated with the first-order approximation, it will be recursively obtained from the initial covariance $\Sigma_{b_{k}}^{b_{k}} = 0$ and the diagonal covariance of noise $Q$.
\begin{figure}[!ht]
\centering
\includegraphics[scale=0.20]{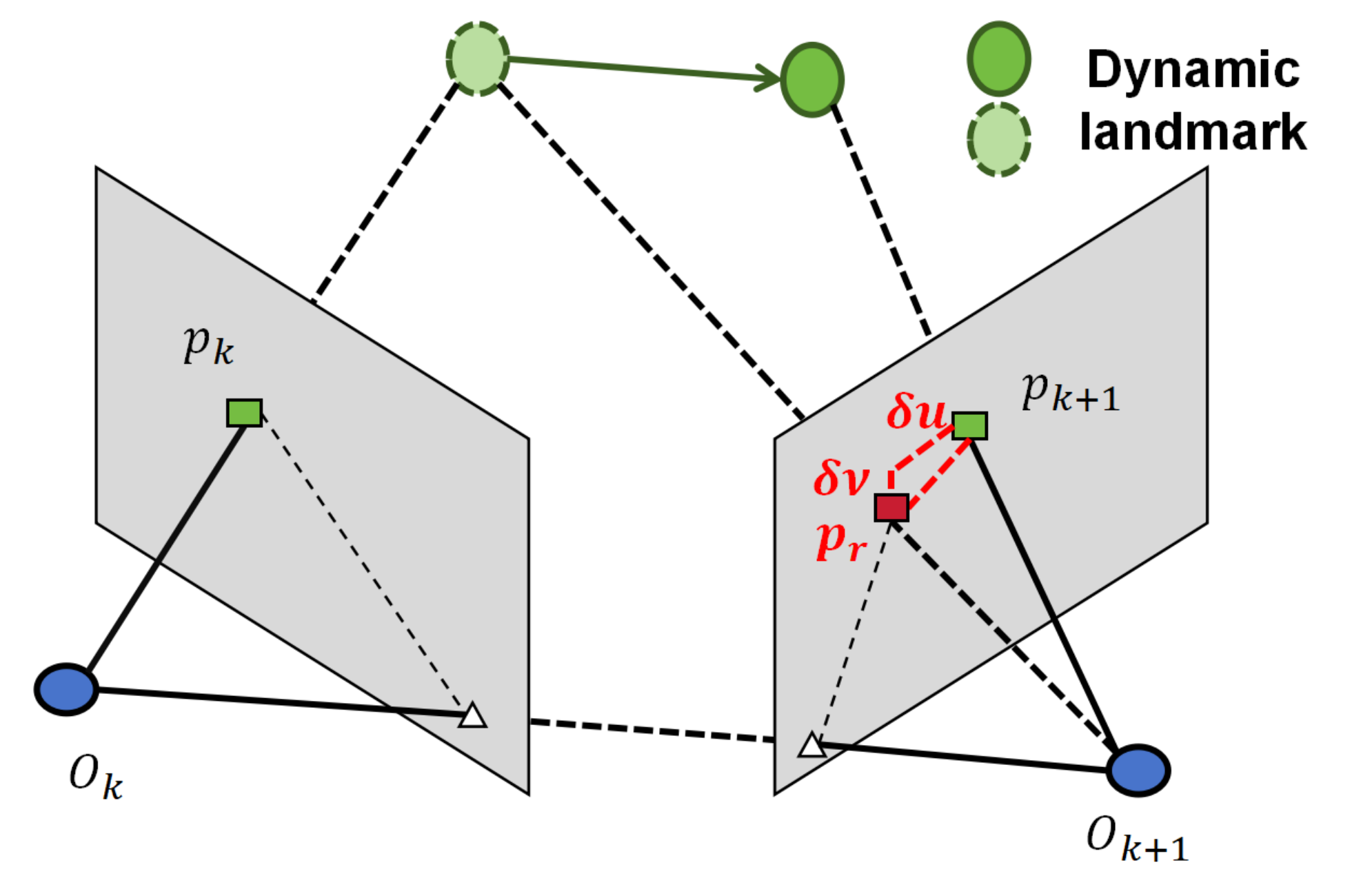}
\caption{\textcolor{black}{\textbf{The epipolar constraint of dynamic landmarks between $kth$ frame and $(k+1)th$ frame.}}}
\label{figurelabel}
\end{figure}
\begin{equation}
    \left\{\begin{array}{c}
    R_{c_{k+1}}^{c_{k}}=R_{c}^{b^{-1}} \cdot R_{b_{k+1}}^{b_{k}} \cdot R_{c}^{b} \quad\text { (a) }\\\\
    t_{c_{k+1}}^{c_{k}}=R_{c}^{b^{-1}} \cdot\left(P_{b_{k+1}}^{b_{k}} -t_{c}^{b}\right) \quad\text { (b) }\\\\
    \begin{aligned}
    \Sigma_{i+1}^{b_{k}} =\left(I+F_{i} \Delta t_{i}\right) \Sigma_{i}^{b_{k}}\left(I+F_{i} \Delta t_{i}\right)^{T} \\
    + \left(G_{i} \Delta t_{i}\right) Q \left(G_{i}\Delta t_{i}\right)^{T}\\
    \text { (c) }
    \end{aligned}
    \end{array}\right\}
\end{equation}
$[R_{c}^{b}$,$t_{c}^{b}]$ represents the extrinsic parameters, $P_{b_{k+1}}^{b_{k}}$ and $R_{b_{k+1}}^{b_{k}}$ respectively are the translation vector and rotation matrix obtained from the propagation in the IMU coordinate system, $b$ denotes the IMU coordinate system while $c$ denotes the camera coordinate system. $i\in[k,k+1]$ represents the IMU between two frames of camera, and $[F_{i},G_{i}]$ is the partial derivative of the residual term and the noise term when the residual is expanded by the first-order Taylor series.
\subsection{Minimum Projection Error and Preprocessing} Considering the epipolar constraint and the projection relationship of the landmarks between two frames of camera as \textbf{Fig.3}, and under the influence of the two-dimensions' projection residual, the precision of the epipolar constraint is not exactly zero.
\begin{equation}
    \begin{aligned}
    e&=x_{k}^{T}\left(t_{c_{k+1}}^{c_{k}}\right)^{\wedge} R_{c_{k+1}}^{c_{k}} x_{k+1}\\
    &=p_{k}^{T} K^{-T}\left(t_{c_{k+1}}^{c_{k}}\right)^{\wedge} R_{c_{k+1}}^{c_{k}} K^{-1} p_{k+1} \\
    &=[a, b, c] \cdot \left[\begin{array}{c}
    u_{k+1} \\
    v_{k+1} \\
    1
    \end{array}\right]=[a, b, c] \cdot \left[\begin{array}{c}
    u_{r}+\delta u \\
    v_{r}+\delta v \\
    1
    \end{array}\right]
    \end{aligned}
\end{equation}
$e$ represents the precision of the epipolar constraint, $[x_{k},x_{k+1}]$ is the coordinate in the normalized camera coordinate system while $[p_{k},p_{k+1}]$ is the coordinate in the pixel coordinate system, $p_{k+1}$ is the coordinate measurement combined the true value $[u_{r},v_{r}]^{T}$ and the residual $[\delta u,\delta v]^{T}$ as \textbf{Fig.3}, $K$ is the internal parameter matrix of camera.

The minimum value of the square projection error is obtained based on the precision of the epipolar constraint and the residual covariance of IMU pre-integration.
\begin{equation}
\left\{\begin{array}{c}
{[a, b, c] \cdot \left[\begin{array}{c}
u_{r} \\
v_{r} \\
1
\end{array}\right]=0,
[a, b, c] \cdot \left[\begin{array}{c}
\delta u \\
\delta v \\
0
\end{array}\right]=e \quad\text { (a) }} \\
\begin{aligned}
    \textcolor{black}{d_{s}=\Sigma^{-1} \min_{\delta u,\delta v} (\delta u^{2} + \delta v^{2}) \quad \text{s.t.} \quad a\delta u + b\delta v = e}\\
\text { (b) }
\end{aligned}\\
\textcolor{black}{\Sigma=\left(1+\sqrt{(\Sigma_{b_{k+1}}^{b_{k}})_{P}} /\left\|t_{c_{k+1}}^{c_{k}}\right\|\right)^{2}}\quad \text { (c) }
\end{array}\right\}
\end{equation}
$d_{s}$ is the minimum square projection error defined by us. $\Sigma^{-1}$ is a factor that characterizes the credibility of the motion prior obtained through IMU propagation, which is defined to avoid the incorrect elimination of dynamic landmarks caused by large IMU propagation error (mainly considered as position error) or the degradation of the epipolar constraint due to an overly small translation amount. \textcolor{black}{$(\Sigma_{b_{k+1}}^{b_{k}})_{P}$ is the variance of the position error of IMU pre-integration between two frames of camera. Noted that for the pinhole camera model, we adopt the projection error on the pixel plane, while for the fisheye camera model, we use the projection unit length error on the normalized plane.}

The square root of projection error $||[\delta u,\delta v]^{T}||$ of landmark tracked between two frames of camera is modeled with a Gaussian distribution. So $\Sigma$ in $\mathbf{(3.c)}$ can be interpreted from the probabilistic perspective as the variance of the Gaussian distribution, $||[\delta u,\delta v]^{T}|| \sim N(0,\Sigma)$. Then the distribution of $d_{s}$ will be approximate to a chi-square distribution with one degree of freedom. The expectation of its coefficient of variation (CV) should approach to $\sqrt{2}$. The elimination and marking strategies for potential dynamic landmarks are as follows.

\textbf{(1) Elimination in preprocess:} In the situation where both dynamic and static landmarks exist, the value of CV is always greater than $\sqrt{2}$. Sort all landmarks in descending order according to the value of $d_{s}$, and start eliminating from the maximum value until the CV no longer decreases or the CV decreases to the target value $\sqrt{2}$. 
 \textcolor{black}{The selected dynamic landmarks are either directly excluded or triangulated to compare the cleanliness of the map.} The elimination criterion and the recursive formulas for mean, variance and CV of remaining landmarks are as follows. 
\begin{equation}
    \left\{\begin{array}{ll}
    \left(d_{s}\right)_{n}>\lambda_{dy} \quad \textit{and} \quad n\in \ell_{dy}, &\textit{dynamic} \\
    \left(d_{s}\right)_{n} \leq \lambda_{dy} \quad \textit{or} \quad n\notin \ell_{dy}, &  \textit{otherwise}
    \end{array}\right.
\end{equation}
\begin{equation}
\left\{\begin{array}{c}
\mu_{n-1}=\left(n \cdot \mu_{n}-\left(d_{s}\right)_{n}\right) /(n-1) \quad\text { (a) }\\
\sigma_{n-1}^{2}=n \cdot \sigma_{n}^{2} /(n-1)-\left(\left(d_{s}\right)_{n}-\mu_{n-1}\right)^{2} / n \text { (b) }\\
C V_{n-1}=\sigma_{n-1} / \mu_{n-1} \quad\text { (c) }
\end{array}\right\}
\end{equation}
$\ell_{dy}$ is the set of landmarks that satisfy above standard. Noted that we implement a strict elimination criterion in order to prevent excessive elimination, only the landmarks with a minimum projection error greater than $\lambda_{dy}$ $pixel^{2}$ will execute the elimination operation. $[\mu_{n},\sigma_{n}^{2}]$ is the mean and variance of the remaining $n$ landmarks.

\textbf{(2) Marking in preprocess:} Ignoring the influence of large IMU propagation error and the degradation of the epipolar constraint, the remaining landmarks with a minimum projection error ($\Sigma\cdot d_{s}$) greater than $\lambda_{dy}^{c}$ $pixel^{2}$ are marked as dynamic candidate landmarks. During the nonlinear optimization process, these landmarks will be treated differently from static landmarks.
\begin{equation}
    \left\{\begin{array}{ll}
    \left(\Sigma\cdot d_{s}\right)_{n}>\lambda_{dy}^{c} \quad , &  \textit{dynamic candidate} \\
    \left(\Sigma\cdot d_{s}\right)_{n} \leq \lambda_{dy}^{c} \quad, & \textit {static}
    \end{array}\right.
\end{equation}

\begin{figure}[!h]
\centering
\subfigure[]{
		\includegraphics[scale=0.15]{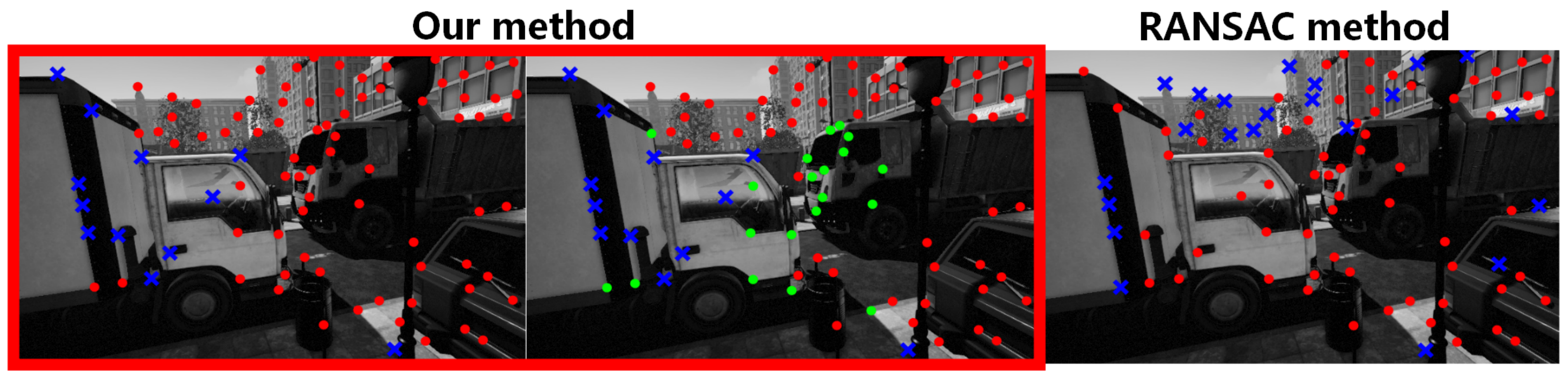}}\\
\centering
\subfigure[]{
		\includegraphics[scale=0.15]{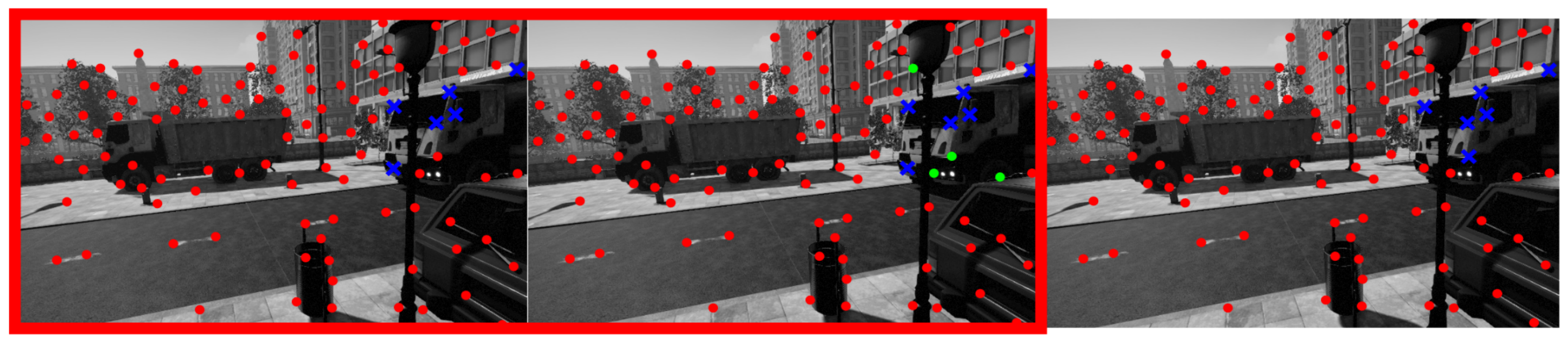}}
\caption{\textbf{Preprocess of dynamic landmarks: }The blue $\color{blue}\times$ marks the rejected dynamic landmarks, the green $\color{green}\bullet$ marks the dynamic candidate landmarks, and the red $\color{red}\bullet$ marks the static landmarks. (a) and (b): The effect comparisons of our method and RANSAC method on the frame with large-scale and small-scale dynamic objects.}
\label{figurelabel}
\end{figure}
\section{ROBUST BUNDLE ADJUSTMENT}
\begin{figure}[!h]
\centering
\includegraphics[scale=0.20]{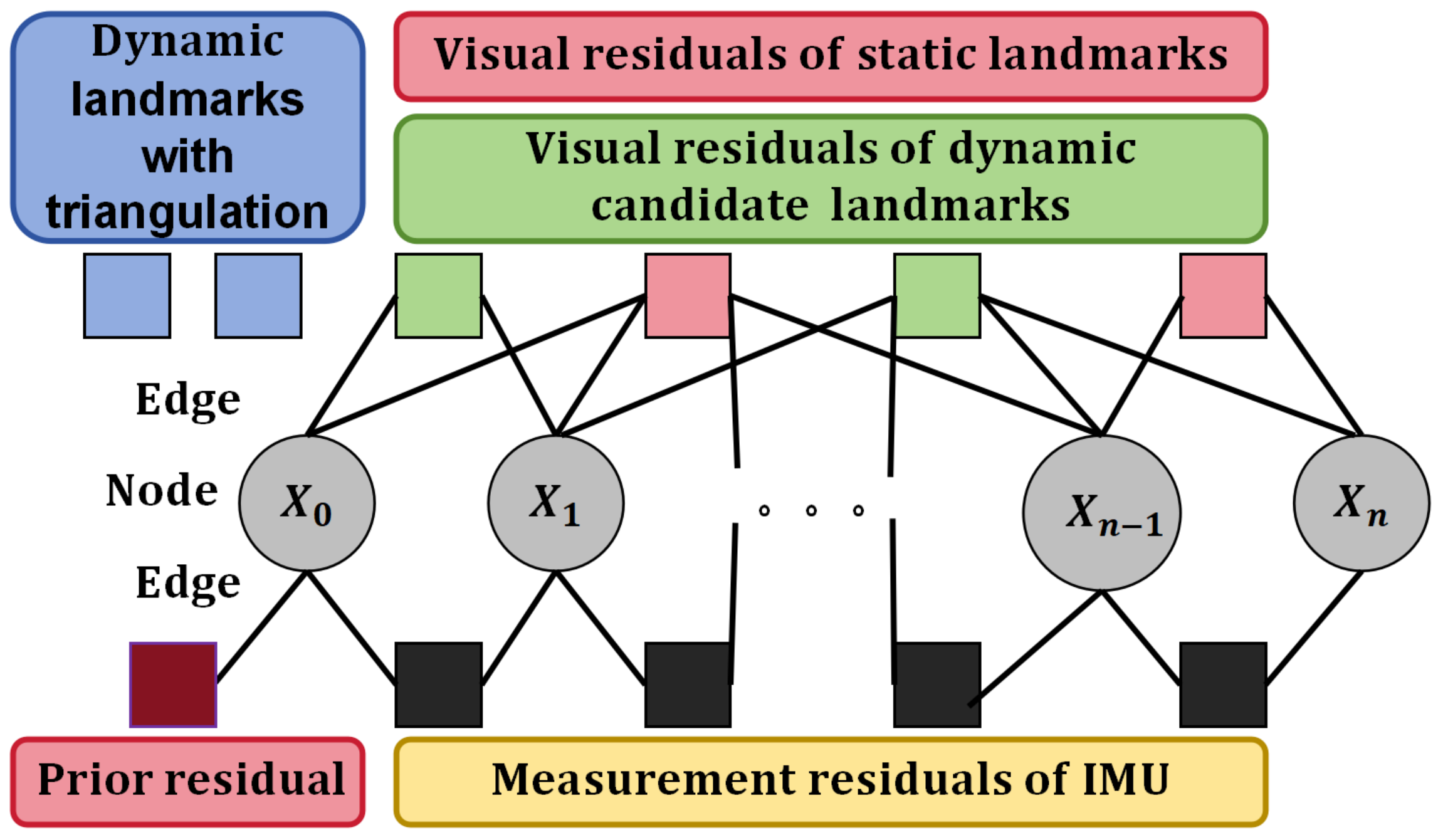}
\caption{\textbf{Sliding window contains static and dynamic candidate landmarks.}}
\label{figurelabel}
\end{figure}
After removing dynamic landmarks and marking dynamic candidate landmarks, tightly coupled \cite{tightly_coupled}, high-precision and robust estimation for the camera poses, landmarks and IMU states in the nonlinear optimization process based on sliding window \cite{vins_mono} is executed with the goal of minimizing a global residual. And the visual residual for dynamic candidate landmarks is proposed in this paper to further reduce the influence of them on the optimization results considering the role of kernel function.

\textcolor{black}{When a dynamic candidate landmark enters the sliding window based optimization process, it is continuously observed for $n$ times. The visual residual for these observations will consider the prior of minimum projection error which will also be updated during the iterative optimization process. Combined with the visual residuals of static landmarks, the measurement residuals of the IMU and the prior residual, the global residual to be minimized is $\textbf{(7)}$. From the probabilistic perspective, the two-dimensions' projection residual of dynamic candidate landmarks satisfies a Gaussian distribution with a covariance matrix of minimum value as $\mathbf{(9)}$.}
\textcolor{black}{\begin{equation}\begin{aligned}
 &\min_{\chi}\left\{
 \left\|r_{p}-H_{p}\chi\right\|^{2}+\sum_{k\in B}\left\|r_{B}(\widehat{Z}_{b_{k+1}}^{b_{k}},\chi)\right\|_{\Sigma_{b_{k+1}}^{b_{k}}}^{2}\right.\\&\left.+\sum_{(s,j)\in\mathcal{C}_s}\rho\left(\left\|r_C(\widehat{Z}_s^{c_j},\chi)\right\|_{\Sigma_v}^2\right)\right.\\&\left.
+\bm{\sum_{(m,j)\in\mathcal{C}_m}\rho\left(\left\|r_C(\widehat{Z}_{m}^{c_j},\chi)\right\|_{\Sigma_{r}^{j}\Sigma_v}^2\right)}
\right\}
\end{aligned}\end{equation}
}
\textcolor{black}{\begin{equation}
\begin{array}{c}
\chi=\left\{{X}_{1}, {X}_{2}, \ldots, {X}_{n}, {X}_{c}\right\} \\
{X}_{{i}}=\left\{{P}_{i}, {V}_{i}, {R}_{{i}}, {b}_{{a}_{{i}}}, {b}_{{g}_{i}}\right\} , i\in[0,1,...,n] \\
{X}_{{c}}=\left\{{\lambda}_{1}, {\lambda}_{2}, \ldots, {\lambda}_{{s}+{m}}\right\}
\end{array}
\end{equation}}
\textcolor{black}{
\begin{equation}
\begin{array}{c}
    \Sigma_{r}^{j}=
    \left[\begin{array}{cc}
    \min\limits_{\delta u_{j},\delta v_{j}} (\delta u_{j}^{2}+\delta v_{j}^{2}) & 0 \\
    0 & \min\limits_{\delta u_{j},\delta v_{j}} (\delta u_{j}^{2}+\delta v_{j}^{2})
    \end{array}\right]\\
    \text{s.t.} \quad a_{j}\delta u_{j} + b_{j}\delta v_{j} = e_{j}
    \end{array}
\end{equation}}
\begin{equation}
    \rho(s) = \left\{\begin{array}{cc}
    s, & s<1 \\
    2 \sqrt{s}-1, & s \geq 1
    \end{array}\right.\\
\end{equation}
\textcolor{black}{${C}_s$, $C_{m}$ and $B$ are respectively the collection of static landmarks, dynamic candidate landmarks and inertial measurements in the sliding window, while $\Sigma_{v}$, $\Sigma_{r}^{j}\Sigma_{v}$ and $\Sigma_{b_{k+1}}^{b_{k}}$ are respectively the covariance matrix of static landmark's projection residual, dynamic candidate landmark's projection residual and IMU pre-integration \cite{imu_preintergration} between the $kth$ and $(k + 1)th$ frames. $[r_{p},H_{p}]$ represents prior information. $\chi$ represents the sum of cameras' state $[{P}_{i}, {V}_{i}, {R}_{i}]$ (position,velocity and rotation at the moment $i$), IMU state $[{b}_{{a}_{{i}}}, {b}_{{g}_{i}}]$ (biases of IMU at the moment $i$) and landmarks' state ${X}_{{c}}$ (the inverse depths of all the landmarks in collection ${C}_s$ and $C_{m}$). $\widehat{Z}_{b_{k+1}}^{b_{k}}$, $\widehat{Z}_s^{c_j}$ and $\widehat{Z}_{m}^{c_j}$ represent the IMU observation, the $jth$ observation of the static landmark, and the $jth$ observation of the dynamic candidate landmark respectively. And $\Sigma_{r}^{j}$ is related to the prior of minimum projection error of the $jth$ observation (the form is similar to formula 3 without considering $\Sigma$) which is set to be $1 pixel^{2}$ when it's value is smaller than $1$) obtained through the epipolar constraint precision between the first observation and the $jth$ $(j\in[2,3,\cdots,n])$ observation of dynamic candidate landmark in the sliding window. $\rho$ is huber kernel function \cite{huber}. Noted that $\left\| *\right\|_{\Sigma }^{2} = \left( *\right)^{T} \Sigma^{-1}\left( *\right)$.}

\begin{figure}[!h]
\centering
\includegraphics[scale=0.4]{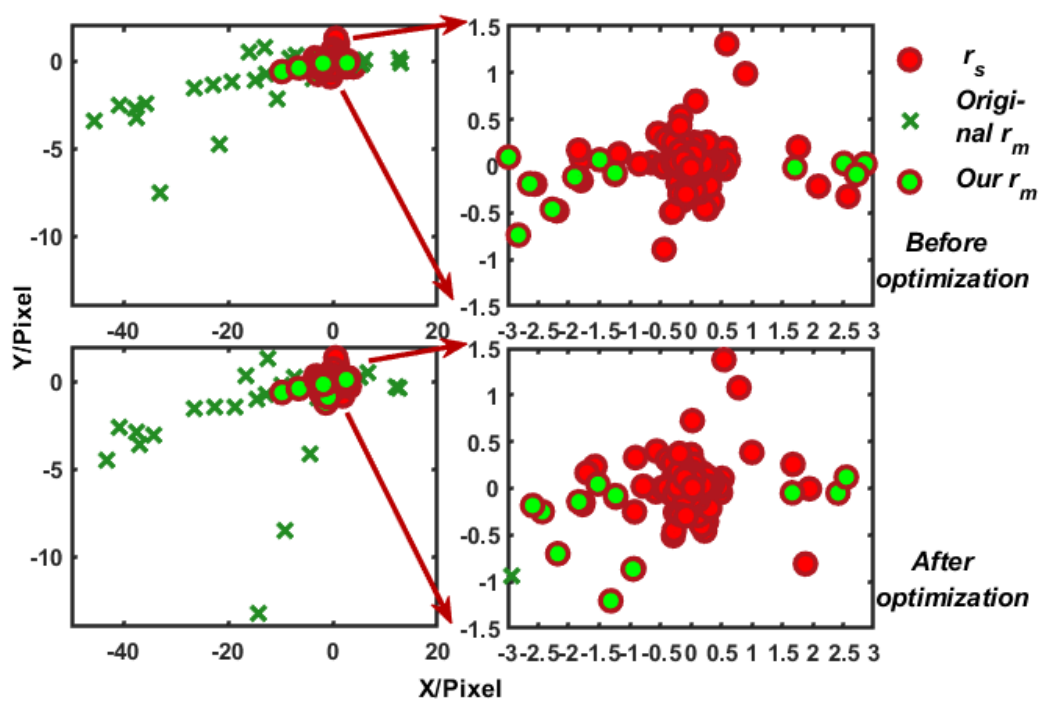}
\caption{\textbf{Comparison between our residuals and original residuals of landmarks: } $r_{s}$ represents residual of static landmarks while $r_{m}$ represents residual of dynamic candidate landmarks.}
\label{figurelabel}
\end{figure}
\textcolor{black}{Fig 6 shows the comparison between the original visual projection residual and the visual projection residual proposed in this paper for dynamic candidate landmarks.} After optimization, 3D-2D projection from the first observation to other relevant observations is performed to calculate the reprojection errors of dynamic candidate landmarks. Remove them outside the sliding window and mark them as confirmed dynamic landmarks if the errors are still relatively large. \textcolor{black}{It should be noted that during the whole process, some unstable landmarks, such as landmarks that are easily occluded, landmarks on the leaves, will also be excluded from the sliding window due to the fact that their tracking results deviate from the motion prior. Finally, a static map which is clean and free of 'ghosting effect' will be obtained.}

\section{EXPERIMENTAL RESULTS}
To evaluate our proposed algorithm, comparisons with the SOTA algorithms in the VIODE and EUROC datasets are carried out, \textcolor{black}{such as VINS-FUSION \cite{vins_fusion}, VINS-Ransac, VINS-Seg and Dyna-VINS \cite{dynavins} in the stereo mode. VINS-Seg is only tested in the VIODE dataset, since it provides instance segmentation templates which can be used to eliminate the landmarks on the dynamic vehicle instances.}
\subsection{Datasets}
\textbf{VIODE:} VIODE dataset \cite{viode} is a simulated dataset that contains lots of dynamic objects, such as cars and trucks. In addition, the dataset includes general occlusion situations. Note that the sub-sequence name none to high means the number and visual field of dynamic objects in the scene.

\textbf{EUROC:} EUROC dataset \cite{euroc} contains different indoor and factory scenes in the real world. There are no dynamic objects but light intensity changes and violent shakings in it. It is used as a verification that our algorithm remains robust for static environments.

\subsection{Remove dynamic landmarks}
After preprocessing to all landmarks during the feature tracking phase, the probabilistic model of landmarks' minimum projection errors is closer to the chi-square distribution with one degree of freedom, and the coefficient of variation (CV) in every frame approaches to $\sqrt{2}$, as \textbf{Fig.7} in VIODE dataset of day sequences from high to none. $\lambda_{dy}$ is set to be \textbf{4} and $\lambda_{dy}^{c}$ is set to be \textbf{1} in experiments.
\begin{figure}[!ht]
\centering
\subfigure[\textbf{Fitted probability density curve}]{
		\includegraphics[scale=0.5]{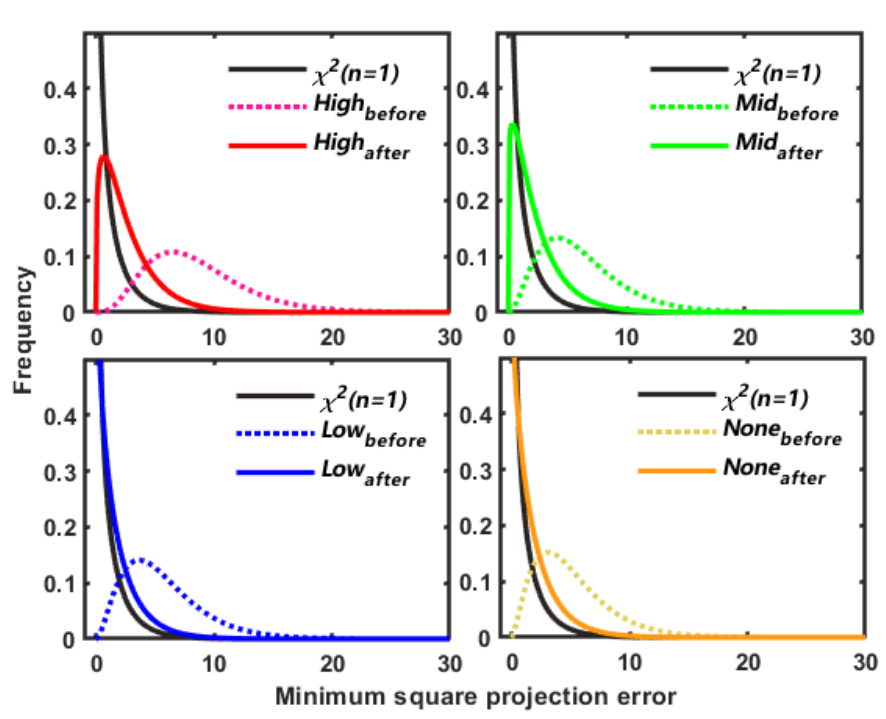}}\\
\centering
\subfigure[\textbf{Coefficient of variation}]{
		\includegraphics[scale=0.5]{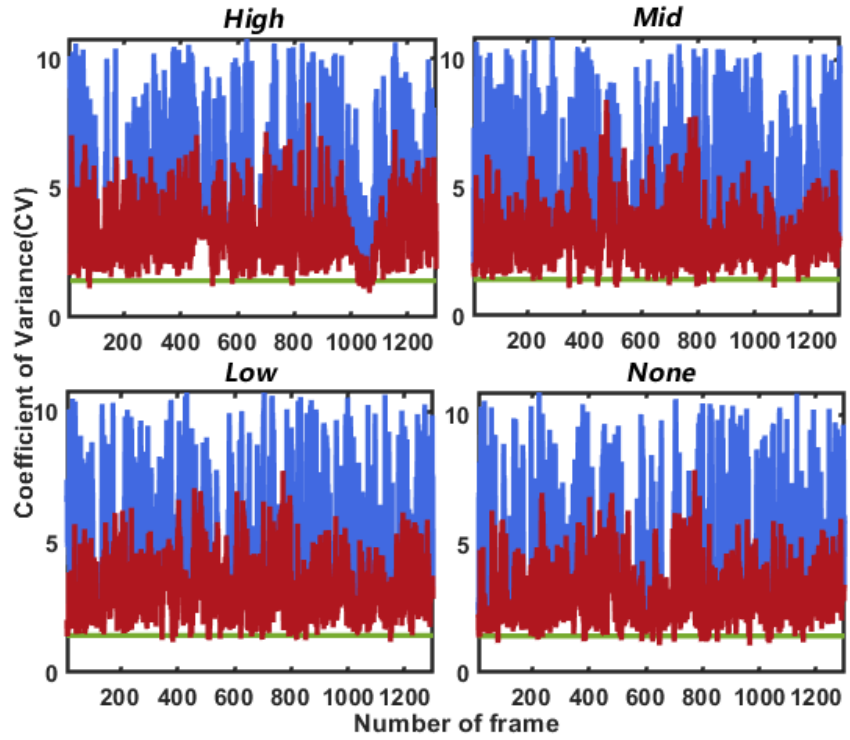}}
\caption{\textbf{Remove dynamic landmarks in preprocess of VIODE day sequences: }(a) : Fitted probability density curve for minimum projection error of sequences from day\_high to day\_none, and the comparison before and after elimination; (b) Coefficient of variation for every frame, blue and red respectively represent CV before and after elimination.}

\label{figurelabel}
\end{figure}
\begin{figure}[!h]
\centering
\subfigure[]{
		\includegraphics[scale=0.15]{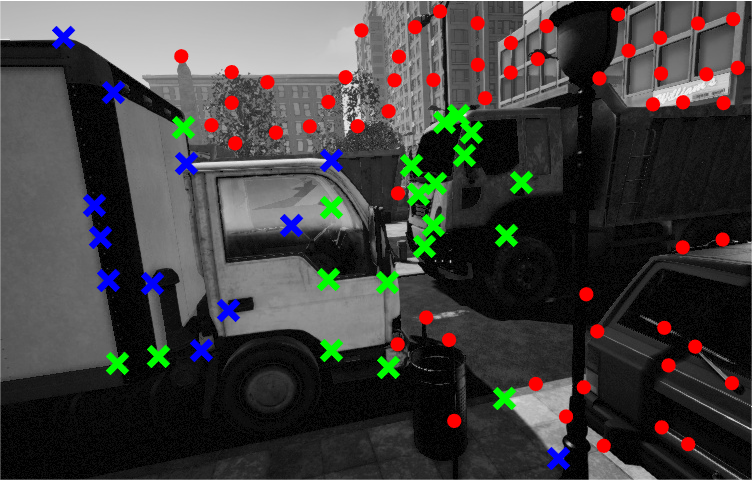}}
\centering
\subfigure[]{
		\includegraphics[scale=0.15]{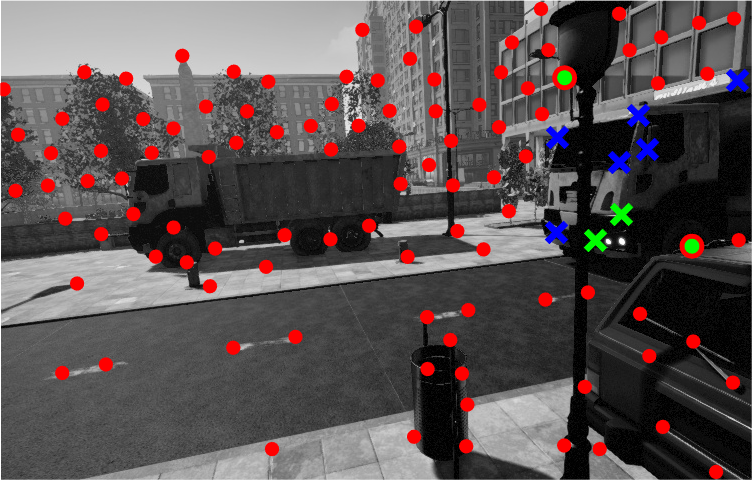}}
\caption{\textbf{Post-process of dynamic landmarks: }The green $\color{green}\times$ marks the confirmed dynamic landmarks, the green $\color{green}\bullet$ in red $\color{red}\circ$ marks the confirmed static landmarks.}
\label{figurelabel}
\end{figure}

After optimization based on sliding window using our visual residual for dynamic candidate landmarks and post-processing to them, confirmed dynamic landmarks will be removed from sliding window while others will be marked as static landmarks. The final labels of landmarks in \textbf{Fig.4} are displayed in \textbf{Fig.8}. \textcolor{black}{And the clean map of static landmarks are displayed in \textbf{Fig.1}.}

\subsection{Quantitative Analysis of Localization}
This subsection provides a quantitative analysis of localization for our method, evaluating the localization accuracy of camera poses and the time taken. Tables I and II present the comparison of our proposed method with other state-of-the-art SLAM methods, using the RMSE (Root Mean Square Error) of ATE (Absolute Trajectory Error) and average tracking and optimization time cost as the evaluation metric. To comprehensively validate the fitness for environments within different dynamic ranges and the effectiveness of the proposed method, we conducted tests on the VIODE dynamic dataset and the EUROC static dataset.
\begin{table*}[!ht]
\begin{center}
\textcolor{black}{\textbf{Table I}~~Comparison of Localization accuracy with state-of-the-art methods (RMSE of ATE in [unit less]). Noted that x represents failure case (diverged).}\\
\setlength{\tabcolsep}{3mm}{
\begin{tabular}{ccccc|cccc|ccc}
\hline
\multirow{2}{*}{Method} & \multicolumn{4}{c}{VIODE city}                                                                                              & \multicolumn{4}{c}{VIODE parking\_lot}                                                                  & \multicolumn{3}{c}{EUROC}                                                            \\ \cline{2-12} 
                        & \multicolumn{1}{l}{none} & \multicolumn{1}{l}{low} & \multicolumn{1}{l}{mid} & \multicolumn{1}{l}{high} & \multicolumn{1}{l}{none} & \multicolumn{1}{l}{low} & \multicolumn{1}{l}{mid} & \multicolumn{1}{l}{high} & \multicolumn{1}{l}{V1\_03} & \multicolumn{1}{l}{V2\_03} & \multicolumn{1}{l}{MH\_03} \\ \hline
VINS-Fusion             & 0.158                         & 0.169                        & 0.193                        & 0.211                         & 0.103                        & 0.126                       & 0.149                       & 0.166                        & 0.216                          & \cellcolor{gray!20}\textbf{0.352}                          & 0.290                          \\
VINS-Ransac      & 0.318                         & 0.289                        & 0.298                        & 0.301                         & \cellcolor{gray!20}\textbf{0.099}                        & \cellcolor{gray!20}\textbf{0.097}                       & x                       & 0.310                        & \cellcolor{gray!40}\textbf{0.109}                          & 0.370                          & \cellcolor{gray!40}\textbf{0.266}                          \\
VINS-Seg      & 0.158                         & 0.152                        & \cellcolor{gray!40}\textbf{0.131}                        &0.247                          & 0.108                        & 0.108                       & \cellcolor{gray!20}\textbf{0.125}                       & 0.140                        & \multicolumn{1}{c}{\textbackslash{}}                         & \multicolumn{1}{c}{\textbackslash{}}                          & \multicolumn{1}{c}{\textbackslash{}}                         \\
Dyna-VINS               & \cellcolor{gray!20}\textbf{0.157}                             & \cellcolor{gray!40}\textbf{0.140}                            & 0.149                            & \cellcolor{gray!40}\textbf{0.122}                             & 0.115                        & 0.099                       & 0.126                      & \cellcolor{gray!20}\textbf{0.124}                        & 0.175                          & 0.479                          & 0.349                          \\
IDY-VINS                &\cellcolor{gray!40}\textbf{0.148}
& \cellcolor{gray!20}\textbf{0.145}                        & \cellcolor{gray!20}\textbf{0.135}                        & \cellcolor{gray!20}\textbf{0.126}                         & \cellcolor{gray!40}\textbf{0.092}                        & \cellcolor{gray!40}\textbf{0.089}                       & \cellcolor{gray!40}\textbf{0.111}                       & \cellcolor{gray!40}\textbf{0.088}                        & \cellcolor{gray!20}\textbf{0.143}                          & \cellcolor{gray!40}\textbf{0.229}                          & \cellcolor{gray!20}\textbf{0.276}                          \\ \hline
\end{tabular}}
\end{center}
\end{table*}

The effects of the proposed method on the localization accuracy is analyzed as shown in Table I. And Table II shows the analysis of our method on the tracking and optimization time cost. \textcolor{black}{The VINS-Seg method does not participate in time test because it's unable to estimate the time required for instance segmentation.}

\textbf{(1) The improvement of localization accuracy for different environments:} According to Table I, our method has a better result than VINS-Fusion and Dyna-VINS. The RANSAC method has the problem of erroneously removing static inliers when facing large-scale dynamic scenes, like \textbf{Fig.4(a)}, which leads to the decline and instability of the accuracy. The method (VINS-Seg) which considers dynamic semantics has not achieved the most ideal localization accuracy results. Especially in the day\_high sequence which contains many scenes with large-area occlusions by dynamic vehicles, it has led to a deterioration in localization. The fact indicates that directly abandon  the constraints of all dynamic landmarks may be too arbitrary. Some dynamic landmarks with slow movement can still provide certainly effective constraints for the system. For city sequences of VIODE from day\_none to day\_high, our method improves the localization accuracy by $\textbf{6.3\%}$, $\textbf{14.2\%}$, $\textbf{30.1\%}$ and $\textbf{40.3\%}$ compared to VINS-Fusion. And our method improves the localization accuracy by $\textbf{10.7\%}$, $\textbf{29.4\%}$, $\textbf{25.5\%}$ and $\textbf{47.0\%}$ compared to VINS-Fusion for indoor parking lot sequences of VIODE from parking\_lot\_none to parking\_lot\_high. The results of EUROC sequences in real world prove that our method still has a stable improvement effect on the localization accuracy in static environments. Moreover, Dyna-VINS method may not be suitable for handling situations such as drastic camera movements and lighting changes in static environments. 

\begin{table*}[!h]
\textbf{Table II}~~Time cost experiment tested in day sequences of VIODE dataset from none to high. Noted that the max number of landmarks every frame is set to 120.

\centering
\setlength{\tabcolsep}{2mm}{
\begin{tabular}{cllllllll}
\hline
\multirow{2}{*}{Method} & \multicolumn{4}{c}{\begin{tabular}[c]{@{}c@{}}Average tracking\\ time cost (ms)\end{tabular}}                                   & \multicolumn{4}{c}{\begin{tabular}[c]{@{}c@{}}Average optimization\\ time cost (ms)\end{tabular}}                           \\ \cline{2-9} 
                        & \multicolumn{1}{c}{day\_none} & \multicolumn{1}{c}{day\_low} & \multicolumn{1}{c}{day\_mid} & \multicolumn{1}{c}{day\_high}     & \multicolumn{1}{c}{day\_none} & \multicolumn{1}{c}{day\_low} & \multicolumn{1}{c}{day\_mid} & \multicolumn{1}{c}{day\_high} \\ \hline
VINS-Fusion             & 19.562                        & 20.671                       & 20.473                       & \multicolumn{1}{l|}{19.568}       & 39.565                        & 38.894                       & 39.245                       & 37.642                        \\
VINS-Ransac     & 19.367+\textbf{0.254}                  & 20.952+0.752                 & 20.727+1.025                 & \multicolumn{1}{l|}{20.053+2.102} & 38.393                        & 38.984                       & 39.848                       & 38.887                        \\
Dyna-VINS               & 19.353                        & 19.504                       & 19.498                       & \multicolumn{1}{l|}{19.703}       & 55.755                        & 48.155                       & 43.628                       & 49.859                        \\
IDY-VINS                & 19.936+0.467                  & 20.397+\textbf{0.362}                 & 20.057+\textbf{0.571}                 & \multicolumn{1}{l|}{19.982+\textbf{0.926}} & 38.375                        & 37.375                       & 37.251                       & 35.564                        \\ \hline
\end{tabular}}
\end{table*}

\textbf{(2) Better time cost performance:} According to Table II, our method does not cause too much additional tracking time cost compared to RANSAC method. However, it should be noted that the time consumption will surely be higher than VINS-Fusion and Dyna-VINS which do not conduct additional elimination operations during the feature tracking phase. On the other hand, the optimized results may converge more quickly for our method than VINS-Fusion and VINS-Fusion\_ransac methods because of the elimination of obvious dynamic landmarks and the classification of dynamic candidate landmarks and static landmarks during optimization process. And the more dynamic objects there are in the environment, the better improvement on the consumption of optimization time will be for our method, while the RANSAC method often requires a longer optimization time to achieve convergence with a lower localization accuracy because of incorrect elimination. Meanwhile, Dyna-VINS method requires much more time for optimization due to its complex processing. The above two aspects of tracking time cost and optimization time cost demonstrate that our method generally has a better real-time performance.

\subsection{\textcolor{black}{Qualitative Analysis of Mapping}}
\textcolor{black}{This subsection provides a qualitative analysis of mapping for our method. Firstly, this paper verifies the recognition and elimination effects of our algorithm on dynamic landmarks under different densities. Taking the day\_high sequence as an example, \textbf{Fig.9} shows the effects of the elimination of dynamic landmarks under low, medium and high mapping densities represented by the landmark densities of 120, 300 and 600 landmarks per frame respectively. The results show that our algorithm has good robustness to different mapping densities. With an increase in mapping density, the recognition and elimination effects of dynamic landmarks and unstable outliers are not attenuated, and relatively clean static mapping results can be obtained.}
\begin{figure*}[!h]
\centering
\subfigure[low]{
		\includegraphics[scale=0.135]{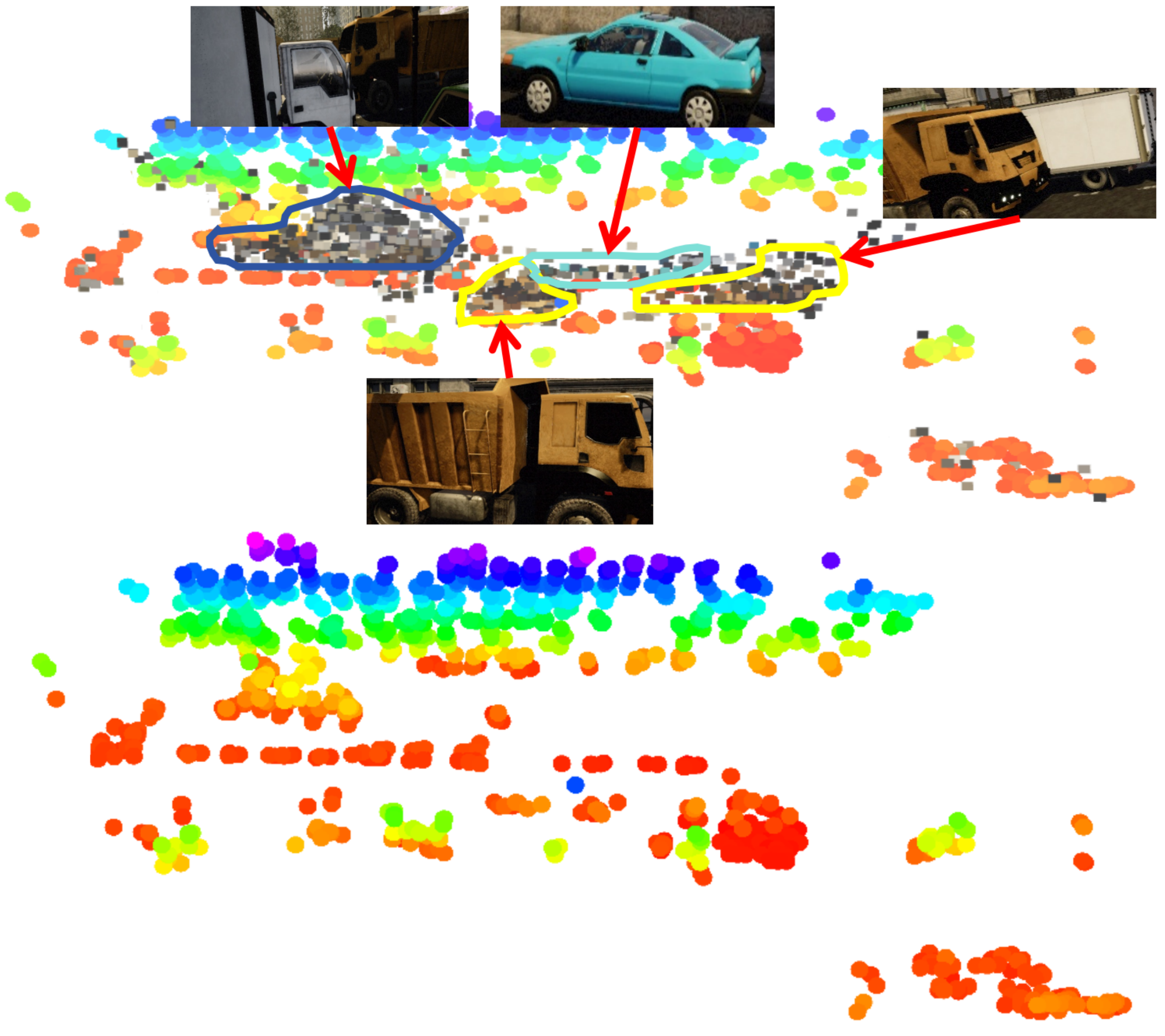}}
\subfigure[medium]{
		\includegraphics[scale=0.135]{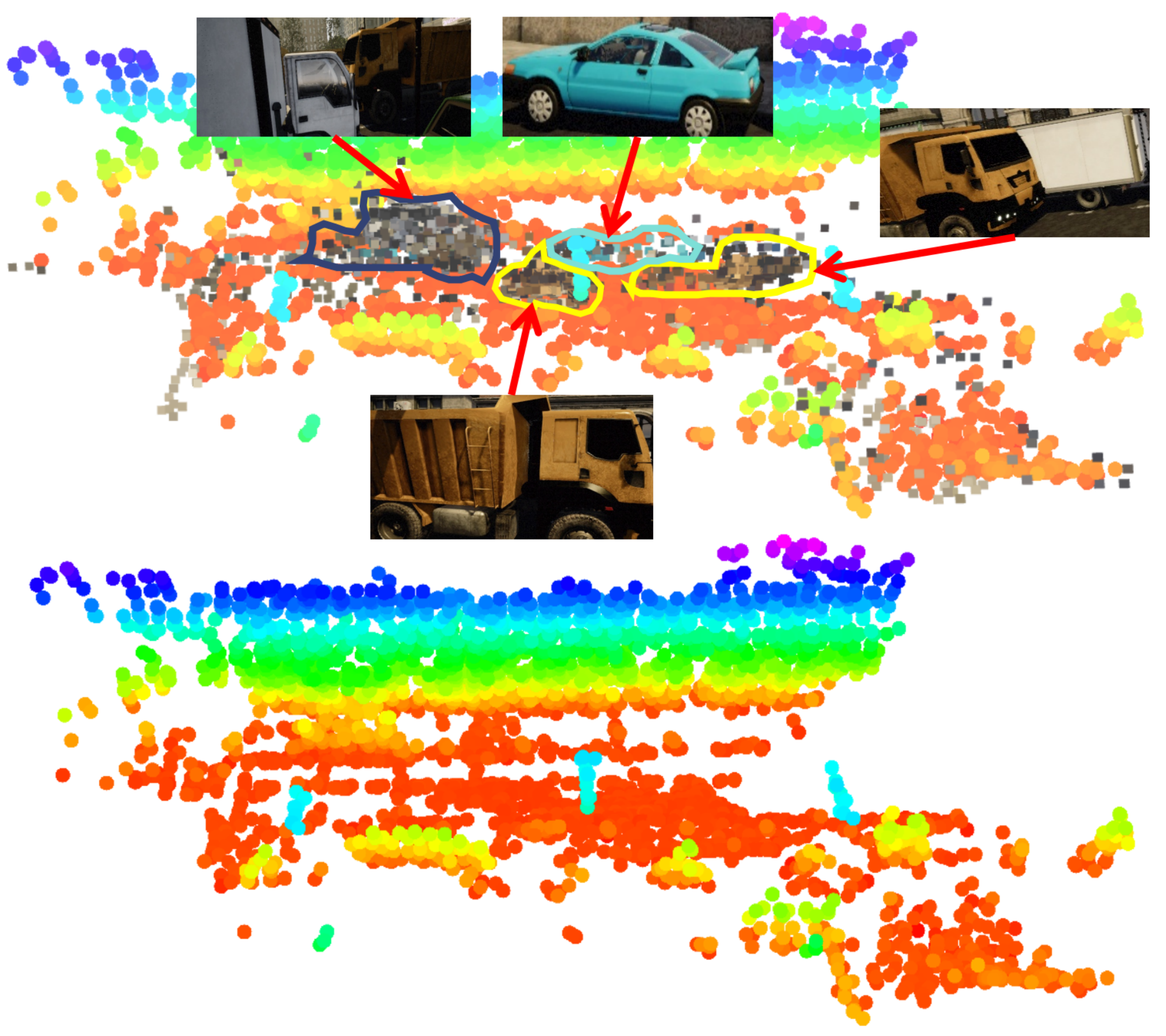}}
\subfigure[high]{
		\includegraphics[scale=0.135]{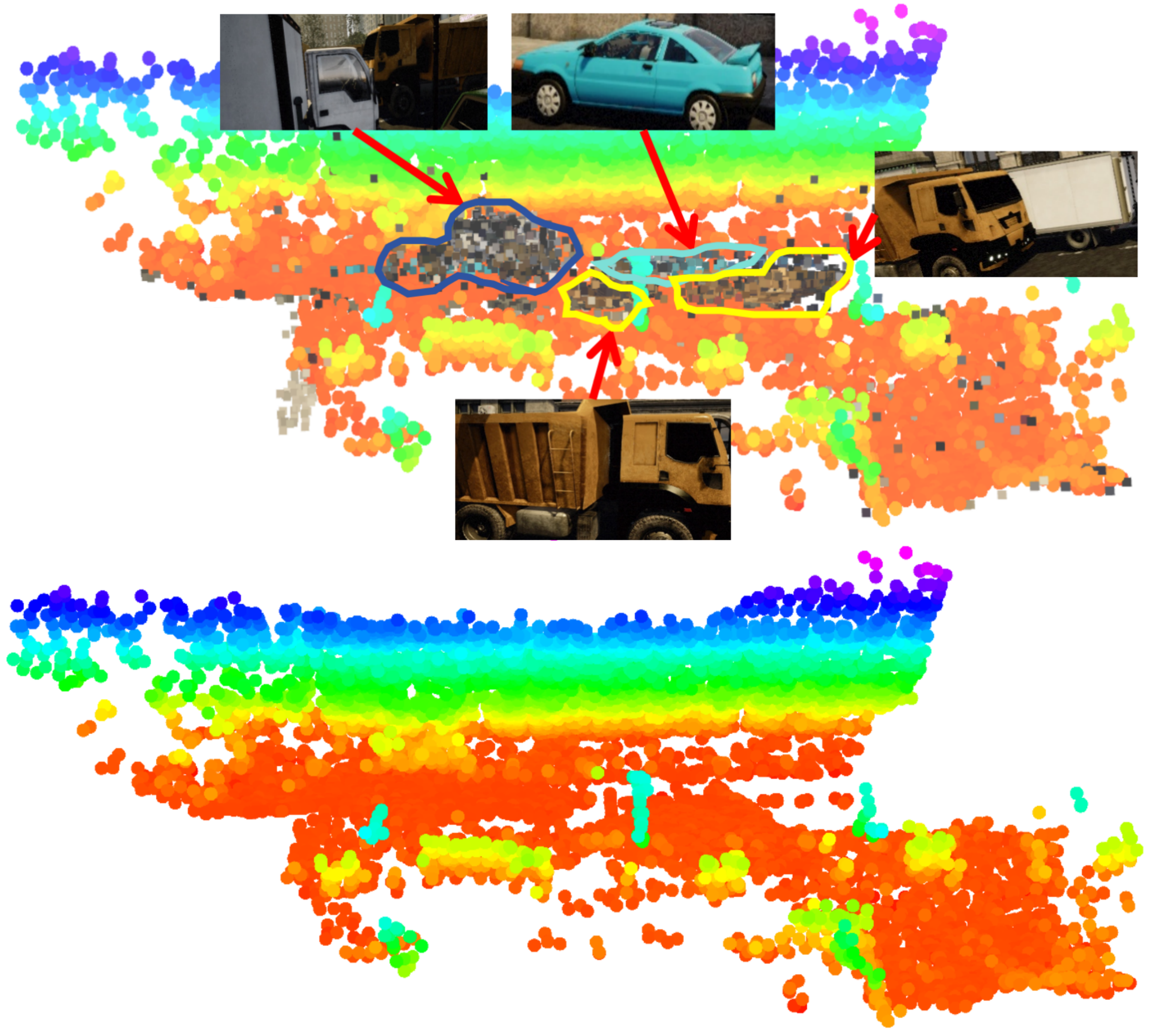}}
\caption{\textcolor{black}{\textbf{Mapping effects of day\_high under different map densities.}}}
\end{figure*}

\textcolor{black}{Secondly, with the same mapping density (600 landmarks per frame for the city sequences and 300 landmarks per frame for the parking\_lot sequences of VIODE), this paper conducted mapping tests on different levels of dynamics from low to high. The results are shown in \textbf{Fig.10}, which verify that our algorithm can effectively recognize dynamic landmarks and leave clean static mapping results under different levels of dynamics, compared to static mapping results of VINS-Seg which abandons all landmarks on dynamic vehicles. The results of VINS-Seg in day\_high and parking\_lot\_high sequences are shown in \textbf{Fig.11}.}

\begin{figure}[!h]
\centering
\subfigure[city sequences (day\_high and static mapping results are shown as Fig.9 (c))]{
		\includegraphics[scale=0.125]{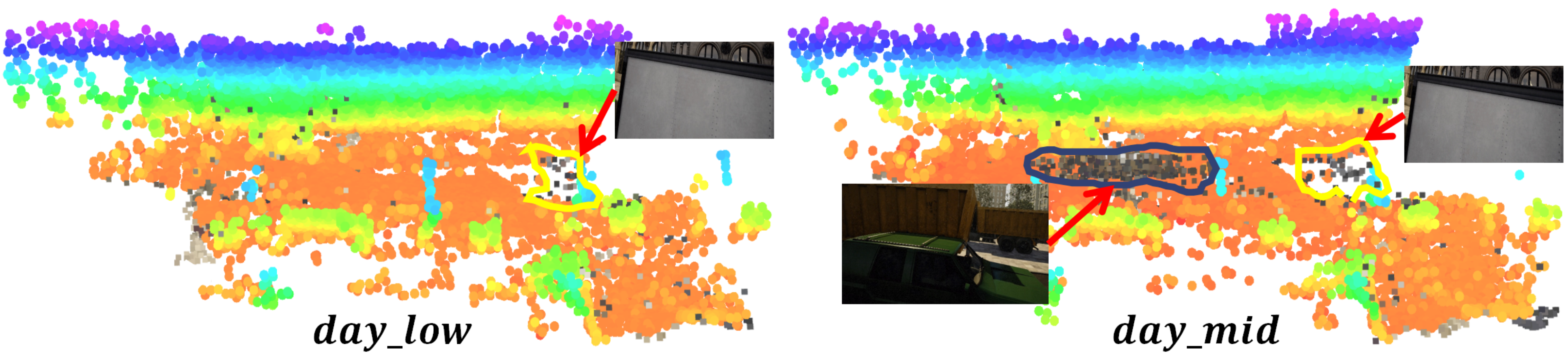}}\\
\subfigure[parking lot sequences]{
		\includegraphics[scale=0.135]{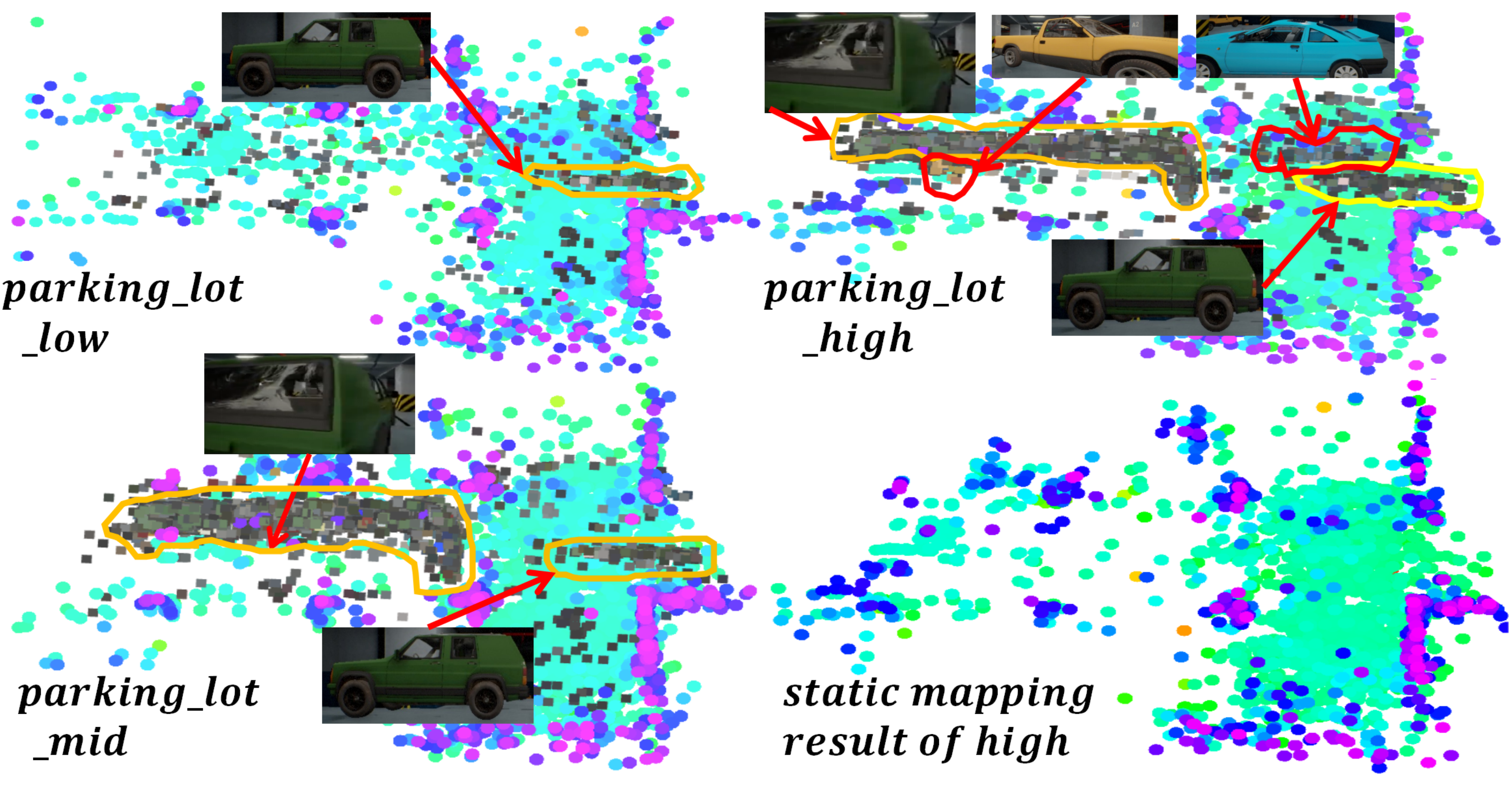}}
\caption{\textcolor{black}{\textbf{Mapping effects of city sequences and parking lot sequences under different levels of dynamics.}}}
\end{figure}
\begin{figure}[!h]
\centering
\includegraphics[scale=0.13]{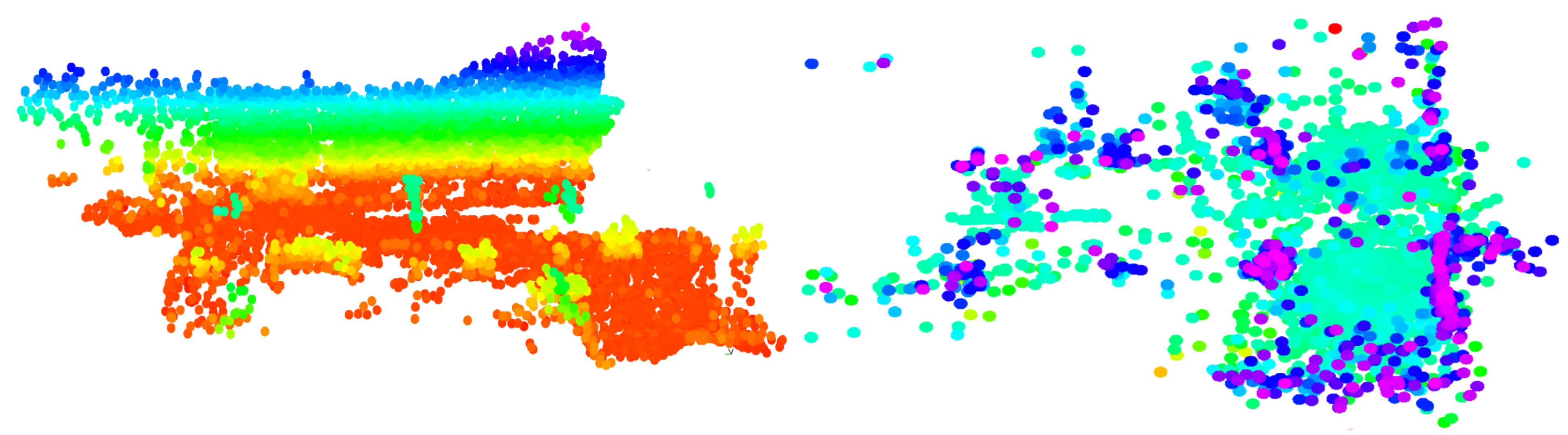}
\caption{\textcolor{black}{\textbf{VINS-Seg's mapping results of day\_high and parking\_lot\_high sequences.}}}
\end{figure}

\section{CONCLUSIONS}
In this paper, the robust visual-inertial motion prior SLAM system (\textit{IDY-VINS}) has been proposed, which effectively handles dynamic landmarks using inertial motion prior for dynamic environments to varying degrees. It is a robust system that preprocesses potential dynamic landmarks and constructs \textcolor{black}{a robust and self-adaptive bundle adjustment residual for dynamic candidate landmarks to reduce their impact. After post-processing, a clean static map is obtained.} Experimental evidences demonstrate that our algorithm has a superior performance compared to other algorithms in both simulated and real environments, both dynamic and static environments.

In future work, \textcolor{black}{we plan to more reasonably arrange the constraints of dynamic landmarks with fully consideration of semantic information and IMU motion prior to improve the system's localization accuracy and mapping effect.} Furthermore, we plan to extend the consideration of dynamic landmarks and candidate landmarks to the overall relocalization and loop closure detection processes of VI-SLAM, \textcolor{black}{so that it can provide a feasible assistance for Life-Long VI-SLAM.}

\addtolength{\textheight}{0cm}   









\bibliographystyle{unsrt}
\bibliography{ref.bib}

\end{document}